# Towards Understanding Egyptian Arabic Dialogues

AbdelRahim A. Elmadany
Department of Science, Institute of Statistical Studies and Research (ISSR), Cairo University

Sherif M. Abdou
Department of Information Technology, Faculty of Computers and Information, Cairo University

Mervat Gheith
Department of Science, Institute of Statistical Studies and Research (ISSR), Cairo University

## ABSTRACT
Labelling of user's utterances to understanding his attends which called Dialogue Act (DA) classification, it is considered the key player for dialogue language understanding layer in automatic dialogue systems. In this paper, we proposed a novel approach to user's utterances labeling for Egyptian spontaneous dialogues and Instant Messages using Machine Learning (ML) approach without relying on any special lexicons, cues, or rules. Due to the lack of Egyptian dialect dialogue corpus, the system evaluated by multi-genre corpus includes 4725 utterances for three domains, which are collected and annotated manually from Egyptian call-centers. The system achieves $F_1$ scores of 70.36% overall domains.

## 1. INTRODUCTION
Knowing the user needs is considered the most important part and the difficult part to build a better human-computer system; which called language understating or somewhere Dialogue Acts (DAs) classification. DAs classification task is labelling the speaker's intention in producing a particular utterance with short words; the DAs terminology is approximately the equivalent of the speech act of Searle (1969) and DAs is different based on dialogue systems domains [1]. Hence, within the field of computational linguistics closely linked to the development and deployment of spoken language dialogue systems, has focused on the some of the more conversational roles such acts can perform. The research on DAs has increased since 1999, after spoken dialogue systems became a commercial reality [2].

This paper refers to an *utterance* as a small unit of speech that corresponds to a single act [3, 4]. In speech research community, *utterance* definition is a slightly different; it refers to a complete unit of speech bounded by the speaker's silence while, we refer to the complete unit of speech as *a turn*. Thus, a single *turn* can be composed of many *utterances*. Turn and *utterance* can be the same definition when the *turn* contains one utterance as defined and used in [5].

There are two approaches to building a DAs classifier: semantic approach and syntax approach. In semantic approach, segment long turns into utterances task is not important because this approach based on turn semantic labelling using identifying the key sequence that called "conceptual segments " or "cues" from the turn. In syntax approach, turn segmentation into utterances task is important segmentation because this approach based morphological features and linguistic rules.

In this paper, we present an approach to understanding spontaneous dialogues for Arabic namely "YOSR". It is a machine learning approach based on context without relying on text diacritization or lexical cues. Whereas, YOSR depends on a set of features from the annotated data that's included morphological features which have been determined by the Morphological Analysis and Disambiguation of Arabic Tool (MADAMIRA)[1] [6], and utterances features. YOSR is evaluated by an Arabic dialogue corpus contains spoken dialogues and Instant Messages (IM) for Egyptian Arabic, and results are compared with manually annotated utterances elaborated by experts.

This paper is organized as follows: section 2 present the challenges of Arabic dialogues understanding, section 3 present the background and related works, section 4 present the proposed classifier "YOSR", section 5 present the experimental setup and results; and finally the conclusion and feature works is reported in section 6.

## 2. ARABIC CHALLENGES
Arabic is one of the six official languages of the United Nations. According to Egyptian Demographic Center, it is the mother tongue of about 300 million people (22 countries). There are about 135.6 million Arabic internet users until $2013^2$.

The orientation of writing is from right to left and the Arabic alphabet consists of 28 letters. The Arabic alphabet can be extended to ninety elements by writing additional shapes, marks, and vowels. Most Arabic words are morphologically derived from a list of roots that are tri, quad, or pent-literal. Most of these roots are tri-literal. Arabic words are classified into three main parts of speech, namely nouns, including adjectives and adverbs, verbs, and particles. In formal writing, Arabic sentences are often delimited by commas and periods. Arabic language has two main forms: Standard Arabic and Dialectal Arabic. Standard Arabic includes Classical Arabic (CA) and Modern Standard Arabic (MSA) while Dialectal Arabic includes all forms of currently spoken Arabic in daily life, including online social interaction and it vary among countries and deviate from the Standard Arabic to some extent [7]. There are six dominant dialects, namely; Egyptian, Moroccan, Levantine, Iraqi, Gulf, and Yemeni.

MSA considered as the standard that commonly used in books, newspapers, news broadcast, formal speeches, movie subtitles… etc. Egyptian dialects commonly known as

---

[1] http://nlp.ldeo.columbia.edu/madamira/
[2] http://www.internetworldstats.com/stats7.htm

Egyptian colloquial language are the most widely understood Arabic dialects due to a thriving Egyptian television and movie industry, and Egypt's highly influential role in the region for much of the 20th century [8]. Egyptian dialect has several large regional varieties such as Delta and Upper Egypt, but the standard Egyptian Arabic is based on the dialect of the Egyptian capital, which is the most understood by all Egyptians.

Due to the lack of an Egyptian Arabic recognition system, manual transcription of the corpus is then required. Moreover, Understanding spontaneous Arabic dialogues task has several challenges:

− Essential characteristics of spontaneous speech: ellipses, anaphora, hesitations, repetitions, repairs… etc. These are some examples from our corpus:
  o A user who does repairs and apologize in his turn: السفر يوم 12 ديسمبر اسفه 11 ديسمبر (*Alsfr ywm 12 dysmbr Asfh 11 dysmbr*, the arrival on 12 sorry 11 December)[3].
  o A user who repeats the negative answer and produce non-necessary information on his turn: لا لا انا مش فاتحة حساب عندكم وانا مبشتغلس بس جوزي هو اللي بيشتغل (*lA lA AnA m$ fAtHp HsAb Endkm wAnA mb$tgls bs jwzy hw Ally by$tgl*, No No I don't have an account in your bank and I'm not an employee but my husband is an employee)

− **Code Switching:** using a dialect words which are derived from foreign languages by code switching between Arabic and other language such as English, France, or Germany. Here an example for user who uses foreign "Egnlish" words in his turn such as ترانزاكشن (*trAnzAk$n*, Transaction) and اكتف (*Aktf*, Active) in his turn. امم فده متاح ولا لازم من الاول اعمل اي ترانزاكشن علشان يبقي اكتف بدل دورمنت (*Ammm fdh mtAH wlA lAzm mn AlAwl AEml Ay trAnzAk$n El$An ybqy Aktf bdl dwrmnt*, Um this is available or I need to do any transaction to activate the dormant account)

− **Deviation:** Dialect Arabic words may be having some deviation such as MSA "اريد" (*Aryd*, want) can be "عايز" (*EAyz*, want), or "عاوز" (*EAwz*, want) in Egyptian dialect.

− **Ambiguity:** Arabic word may be having different means such as the word "علم" can be: "عَلَمْ" "*flag*", "عِلْمْ" "*science*", "عُلِمَ" "*it was known*", "عَلِمَ" "*he knew*", "عَلَّمَ" "*he taught*" or "عُلِّمَ" "*he was taught*". Thus, the ambiguity considers the key problem for Natural Language Understanding/Processing especially on the Arabic language. The word diacritization is very useful to clarify the meaning of words and disambiguate any vague spellings.

− **Lack of Resources:** The not existence and the lack of tagged Arabic Spontaneous Dialogues and Instant Messages corpora for Egyptian Arabic corpus make turn segmentation task far more challenging. Since manual construction of tagged corpus is time-consuming and expensive [9], it is difficult to build large tagged corpus for Arabic dialogue acts. Therefore, the researchers had to build their own resources for testing their approaches. Consequently, we used JANA corpus, which a multi-genre corpus of Arabic dialogues labeled for Arabic Dialogues Language Understanding (ADLU) at utterance level ant it comprises spontaneous dialogues and IM for Egyptian dialect; it developed by [10, 11]. JANA corpus has collected, segmented and annotated manually. JANA consists of approximately 3001 turns with average 6.7 words per turn, contains 4725 utterances with average 4.3 words per utterance, and 20311 words. The list of dialogue acts "dialogue acts schema" which have been used in the annotation of utterances in JANA corpus is shown in Table 1; this schema developed by [12].

**Table 1. Dialgoues Acts Annotation Schema**

| Request Acts | Response Acts |
|---|---|
| Taking-Request | Service-Answer |
| Service-Question | Other-Answer |
| Confirm-Question | Agree |
| YesNo-Question | Disagree |
| Choice-Question | Greeting |
| Other-Question | Inform |
| Turn-Assign | Thanking |
| **Other Acts** | Apology |
| Opening | MissUnderstandingSign |
| Closing | Correct |
| Self-Introduce | Pausing |
|  | Suggest |
|  | Promise |
|  | Warning |
|  | Offer |

## 3. RELATED WORKS

Webb and Hardy are noticing that there are two ways to understand the dialogues language [13]:

− **Shallow understanding:** It is simple spotting keywords, or having lists of, for example, every location recognized by the system. Several systems are able to decode directly from the acoustic signal into semantic concepts precisely because the speech recognizer already has access to this information.

− **Deeper analysis:** Using linguistic methods; including part-of-speech (POS) tagging, syntactic parsing and verb dependency relationships.

Using Machine Learning (ML) for solving the DA classification problem, researchers have not historically published the split of training and testing data used in their experiments, and in some cases methods to reduce the impact of the variations that can be observed when choosing data for training and testing have not been used [3]. Moreover, DAs are practically used in many live dialogue systems such as Airline Travel Information Systems (ATIS) [14], DARPA [15], and VERBMOBIL project [16]. N-gram models can be considered the simplest method of DA classification based on some limited sequence of previous DAs such as [3, 13, 17, 18] and sometimes used with Hidden Markova Model (HMM) such as [19]. In addition, there are other approaches such as Transformation-Based Learning (TBL)[20], and Naïve Bayesian [21]. These approaches are tested and designed for non-Arabic dialogues such as English, Germany, and France that completely differs for Arabic dialogues.

To the best of our knowledge, there are few works interested in Arabic dialogue acts classification such as [22] are used Naïve Bayes and Decision Trees. [23] are used utterances semantic labelling based on the frame grammar formalism. [24] are used syntactic parser context free grammar with HHM. [5] are Conditional Random Fields (CRF) to semantically label spoken Tunisian dialect turns. [25, 26] are used Arabic function words such as "هل" "do/does", "كيف"

---

[3] Examples are written as Arabic (*Buckwalter transliteration schema*, English translation)

"How" to classify questions and non-questions utterances with Decision Tree classifier. These approaches designed and applied on MSA or Tunisian dialect. To the best of our knowledge, there is no published word for understanding Egyptian Arabic or Egyptian dialect. The survey by [27] presents background and the progress made in understanding Arabic dialogues.

In this work we complete the work of [28] which present the first steps for understanding the Egyptian Arabic when built the annotated corpus for Egyptian Arabic dialogues namely "JANA" and present the turns segmentation into utterances classifier.

## 4. METHODOLOGY

Support Vector Machine (SVM) is a supervised machine learning that has been shown to perform well on text classification tasks, where data is represented in a high dimensional space using sparse feature vectors [29, 30]. Moreover, the SVM is robust to noise and the ability to deal with a large number of features effectively [31]. The SVM classifier is trained to discriminate between examples of each class, and those belonging to all other classes combined. During testing, the classifier scores on an example, are combined to predict its class label [32].

Most of works have dealt with the dialogues act classification problem for spoken dialogues using SVM; they usually used a semantic labelling. In this paper, we proposed a classifier based on multi-class SVM that reduces the running time, and reduce the training cost and time. For instance, if we have 24 dialogues acts, we needs 24 binary classifiers, but here we proposed a one classifier. We are working on the hypothesis that the dialogues act problem can solved as text-chunking problem.

The proposed approach "YOSR" is an SVM approach, which a Machine Learning based, involves a selected set of features, extracted from annotated utterances, that is used to generate a statistical model for utterance act prediction. We used YamCha SVM toolkit[4] that converts utterances classification task to a text-chunking task.

There are three processes to do as preprocessing the input utterances before running the YOSR classifier. These processes are:

– **Normalization:** to avoid writing errors from the transcription, we normalized the transcribed turns (unified Arabic characters) as
  o Convert Hamza-under-Alif "إ", Hamza-over-Alif "أ", and Madda-over-Alif "آ" to Alif "ا"
  o Convert Teh-Marbuta "ة" to Heh "ه"
  o Convert Alif-Maksura "ى" to Yeh "ي".
– **Split "و" (w, and) from the original words:** Sometimes the writers write the conjunction "و" (w, and) concatenated to the next word. For instant, "وقال" (wqAl, and he talked) the original word "قال" (qAl, he talked) is concatenated with the conjunction "و" (w, and). Detect and split "و" (w, and) from the original words has been done using *Wawanizer Toolkit* [10].
– The utterances are transliterated from Arabic to Latin based ASCII characters using the Buckwalter transliteration scheme[5].

---
[4] Available at http://chasen.org/~taku/software/yamcha/
[5] http://languagelog.ldc.upenn.edu/myl/ldc/morph/buckwalter.html

There are two phases has employed for carrying out the classification task in our approach, training phase and test phase. The training phase generates the classification model using a set of classification features. In the test phase, the classification model is utilized to predict a class for each token (word). In the training phase, each word is represented by a set of features and its actual DA tag in order to produce an SVM model that is able to predict the dialogue act of the utterance.

We used the BIO format (Beginning *B* of the ACT, Inside *I* the ACT, and Outside *O* the ACT) or sometimes IOB Format, which developed for text-chunking by [33]. For instance, each word in utterances that refer to dialogue act "*Service-Question*" will be represented as the first word of utterance will tagged by "*B- Service_Question*" and the other words will tagged by "*I-Service_Question*".

The first step in our approach is to extract the significant features from the training data. Consequently, we study the impact of the features individually by using only one feature at a time and measure the classifier's performance using the F-measure metric. Finally, according to the performance achieved, we select the optimized features for the proposed approach "YOSR".

### 4.1. Features Selection

Feature selection refers to the task of identifying a useful subset of features chosen to represent elements of a larger set.

**Contextual word:** The features of a sliding window, including a word n-gram that includes the candidate word, along with previous and following words. For instance, in the training corpus the word "عايز" (*EAyz*, want) appears frequently before a user's request that indicate a *Request* act or *Service-Question* act. Therefore, the classifier will use this information to predict a *Service-Question* act for this utterance.

**Morphological Features:** We used word Part-Od-Speech (POS); the sequence of NOUN and PROPER NOUN indicates the speaker needs to introduce himself or his company, for example "مصر للطيران" (*mSr llTyrAn*, Egypt Air). In addition, the sequence of preposition, NOUN or PROPER NOUN indicates the speaker needs to greet or return greet the other one. The consequence of NOUN and PROPER NOUN indicate the speaker needs to introduce himself or his company, for example "وعليكم السلام" (*wElykm AlslAm*, Peace be upon you).

**Utterances Features:** Using of the utterances meta information can help dialogue act classification process [34-36]. Also, Knowing what happened before current utterance can help the classification task [37, 38]. We used:

– *Utterance Speaker Type:* the speaker type Operator or Customer of the current utterance can help to determine the act of utterance. For instance, the act "Service-Question" is related to the customer because he connected to service support service to asking about a provided service, but the act "Other-Question" and "Choice-Question" is related to operator because the operator asking the client for his name or choosing the client to select one of the provided services.
– *Previous Utterance Act*: Knowing the previous utterances acts sequence in the dialogue help the classifier to predict the act of current utterance. For instance, the act "Agree" and "Disagree" is almost followed by the "Confirm-Question" act.

## 5. EXPERIMENTAL SETUP AND RESULTS

In order to measure the effect of complexity of each dialogues domain (Banks, Flights, and Mobile Network Operators) on classification accuracy, we experiment on each dialogue domain separately and one experiment to overall combined data. We split each domain based on dialogue turn boundary into 70% training dataset, 20% development dataset (DEV), and the 10 % test dataset as shown in Table 2. The results are reported using standard metrics of Accuracy (Acc), Precision (P), Recall (R), and the F1 score ($F_1$).

$$F_1 = \frac{2PR}{P+R} \quad (1)$$

**Table 2. Corpus training, development (DEV) and test datasets**

| | Domain | Datasets | Dialogues | Turns | Utterances |
|---|---|---|---|---|---|
| Spoken | Banks | DEV | 4 | 115 | 193 |
| | | Test | 5 | 226 | 368 |
| | | Training | 17 | 782 | 1,234 |
| | Flights | DEV | 5 | 145 | 242 |
| | | Test | 7 | 224 | 364 |
| | | Training | 14 | 773 | 1,186 |
| IM | Mobile Network Operators | DEV | 3 | 75 | 109 |
| | | Test | 5 | 197 | 272 |
| | | Training | 22 | 464 | 757 |
| | **Total** | | **82** | **3,001** | **4,725** |

In the training stage, the training is applied on the training dataset using selected features set and the results are analyzed to determine the best features set. The development stage is performed using the DEV dataset to define the best feature set which used in the test stage. In the test stage, the classifier is applied on the test dataset and the results are reported and discussed.

**Table 3. The results of Banks test set**

| Act | Precision | Recall | $F_1$ |
|---|---|---|---|
| Agree | 90.57 | 97.96 | 94.12 |
| Closing | 100 | 100 | 100 |
| Confirm_Question | 32.56 | 45.16 | 37.84 |
| Disagree | 100 | 60 | 75 |
| Greeting | 92.86 | 96.3 | 94.55 |
| MissUnderstandingSign | 60 | 75 | 66.67 |
| Other_Answer | 46.15 | 40 | 42.86 |
| Other_Question | 71.43 | 38.46 | 50 |
| Pausing | 50 | 25 | 33.33 |
| SelfIntroduce | 100 | 30 | 46.15 |
| Service_Answer | 69.07 | 81.71 | 74.86 |
| Service_Question | 78.26 | 56.25 | 65.45 |
| Taking_Request | 100 | 100 | 100 |
| Thanking | 69.23 | 81.82 | 75 |
| Turn_Assgin | 72.73 | 100 | 84.21 |
| **Over All** | **72.75** | **72.75** | **72.75** |

We classify utterances labelling task as a Multi-classification problem. Therefore, the proposed approach is tested using PAIRWISE and ONE vs ALL (OVA) approaches and we found the ONE vs ALL approach achieves the best performance in this task. Moreover, the selected features are tested on window size within ranges from -1/+1 to -5/+5. We found that a context size of -2/+2 achieves the best performance.

**Table 4. The results of Flights test set**

| Act | Precision | Recall | F1 |
|---|---|---|---|
| Agree | 93.42 | 84.52 | 88.75 |
| Closing | 100 | 100 | 100 |
| Confirm_Question | 44.44 | 26.09 | 32.88 |
| Disagree | 33.33 | 18.18 | 23.53 |
| Greeting | 90.91 | 88.24 | 89.55 |
| Other_Answer | 29.17 | 58.33 | 38.89 |
| Other_Question | 37.5 | 50 | 42.86 |
| Pausing | 25 | 16.67 | 20 |
| SelfIntroduce | 92.86 | 100 | 96.3 |
| Service_Answer | 59.77 | 71.23 | 65 |
| Service_Question | 48.89 | 64.71 | 55.7 |
| Thanking | 70 | 63.64 | 66.67 |
| Turn_Assgin | 66.67 | 57.14 | 61.54 |
| **Over All** | **65** | **64.11** | **64.55** |

**Table 5. The results of Mobile Network Operators test set**

| Act | Precision | Recall | F1 |
|---|---|---|---|
| Agree | 65.52 | 63.33 | 64.41 |
| Apology | 100 | 25 | 40 |
| Confirm_Question | 66.67 | 58.33 | 62.22 |
| Disagree | 50 | 60 | 54.55 |
| Greeting | 91.3 | 84 | 87.5 |
| Other_Question | 81.82 | 90 | 85.71 |
| Pausing | 100 | 63.64 | 77.78 |
| SelfIntroduce | 100 | 76.92 | 86.96 |
| Service_Answer | 64.29 | 88.52 | 74.48 |
| Service_Question | 58.33 | 80 | 67.47 |
| Thanking | 90.91 | 83.33 | 86.96 |
| Turn_Assgin | 60 | 81.82 | 69.23 |
| **Over All** | **68.01** | **68.27** | **68.14** |

Table 3, Table 4 and Table 5 shows the results for each domain Banks, Flights, and Mobile Networks Operators respectively and Table 6 shows the results of overall combined data experiment. The results show that Flights dialogues have $F_1$ fewer than the other domains. Moreover, the analysis of error results shows that the errors occur due to one of these reasons:

− There are some sentences/words can refers to more than meaning "dialogue acts" such as:

  o "شكرا" (*$krA*, Thank you) usually it means "thanks" (thank you) but sometimes it means "disagree" (No) when comes after utterance such as "أي استفسار تاني حضرتك" (*>y AstfsAr tAny HDrtk*, Any other services)

  o "عفوا" (*EfwA*, you are welcome) usually it means "you are welcome" as a reply of "thank you" statement, but sometimes it means "miss understanding sign" (sorry).

− Some utterance needs to add semantic features to the classifier "deeply semantic analysis". For instance, for the operator's utterance such as "ولكن طبعا لازم يكون عدي عليها 6 شهور" (*wlkn TbEA lAzm ykwn Edy ElyhA 6 $hwr*, Make sure you must get it since 6 months). The customer has responded utterance such as لا لا هي عدي عليها 4 سنين (*lA lA hy Edy ElyhA 4 snyn*, No it since 4 years). The proposed classifier is classified the customer utterance as "disagree

act" because it contains "لا" (*lA*, No) in spite of the customer agreement with the operator warning.

– The instant messages utterances (Mobile Network Operators) are contained many of writing errors that confused the classifier

**Table 6. The results of Overall combined test set**

| Act | Precision | Recall | F1 |
|---|---|---|---|
| Agree | 86.94 | 91.04 | 88.94 |
| Apology | 100 | 50 | 66.67 |
| Closing | 100 | 100 | 100 |
| Confirm_Question | 40 | 33.66 | 36.56 |
| Disagree | 76.47 | 50 | 60.47 |
| Greeting | 91.57 | 88.37 | 89.94 |
| MissUnderstandingSign | 100 | 20 | 33.33 |
| Other_Answer | 39.02 | 47.06 | 42.67 |
| Other_Question | 58.33 | 60 | 59.15 |
| Pausing | 71.43 | 47.62 | 57.14 |
| SelfIntroduce | 96.43 | 75 | 84.37 |
| Service_Answer | 67.44 | 80.56 | 73.42 |
| Service_Question | 60.83 | 72.28 | 66.06 |
| Taking_Request | 100 | 75 | 85.71 |
| Thanking | 77.14 | 79.41 | 78.26 |
| Turn_Assgin | 76.67 | 88.46 | 82.14 |
| **Over All** | **70.61** | **70.12** | **70.36** |

Moreover, the results show that YOSR classifier yields a good performance and efficiency in understanding Arabic Egyptian dialect dialogues for all domains without using any special lexicons, cues, or rules for each domain.

## 6. CONCLUSION

In this paper, we present a ML approach using SVM to solve the problem of automatic understanding of the Arabic dialogues task for Egyptian dialect at the utterance level; namely, YOSR. The proposed classifier has tested on corpus consists of spontaneous dialogues and IM for Egyptian dialect.

The results obtained that YOSR classifier is very promising. To the best of our knowledge, these are the first results reported for understanding the Egyptian dialect.

As perspectives, we plan to improve YOSR by adding a general cues for call-centers domain, morphological features such as the first verb type and Lemma, context-based features, and dialect words treatments. Moreover, we intend to extend the training corpus "JANA" to improve the classification results.